\documentclass{article}

\usepackage{microtype}
\usepackage{graphicx}
\usepackage{subfigure}
\usepackage{booktabs} % for professional tables

\usepackage{hyperref}

\usepackage{enumitem}
\usepackage{amsmath}
\usepackage{wrapfig}
\usepackage{caption}
\usepackage{mathtools}
\usepackage{amsfonts}

\newtheorem{remark}{Remark}
\newtheorem{theorem}{Theorem}

\newcommand{\argmax}{\operatornamewithlimits{argmax}}

\usepackage[accepted]{icml2021}

\icmltitlerunning{Class2Simi: A Noise Reduction Perspective on Learning with Noisy Labels}

\begin{document}

\twocolumn[
\icmltitle{Class2Simi: A Noise Reduction Perspective on Learning with Noisy Labels}



\icmlsetsymbol{equal}{*}

\begin{icmlauthorlist}
\icmlauthor{Songhua Wu}{equal,usyd}
\icmlauthor{Xiaobo Xia}{equal,usyd}
\icmlauthor{Tongliang Liu}{usyd} \\
\icmlauthor{Bo Han}{hkbu} 
\icmlauthor{Mingming Gong}{mel}
\icmlauthor{Nannan Wang}{xidian}
\icmlauthor{Haifeng Liu}{leinao}
\icmlauthor{Gang Niu}{riken}
\end{icmlauthorlist}

\icmlaffiliation{usyd}{Trustworthy Machine Learning Lab, School of Computer Science, The University of Sydney}
\icmlaffiliation{hkbu}{Department of Computer Science, Hong Kong Baptist University}
\icmlaffiliation{riken}{RIKEN Center for Advanced Intelligence Project}
\icmlaffiliation{mel}{School of Mathematics and Statistics, The University of Melbourne}
\icmlaffiliation{xidian}{ISN State Key Laboratory, School of Telecommunications Engineering, Xidian University}
\icmlaffiliation{leinao}{Brain-Inspired Technology Co., Ltd.}

\icmlcorrespondingauthor{Tongliang Liu}{tongliang.liu@sydney.edu.au}

\icmlkeywords{Machine Learning, ICML}

\vskip 0.3in
]



\printAffiliationsAndNotice{\icmlEqualContribution} % otherwise use the standard text.

\begin{abstract}
Learning with noisy labels has attracted a lot of attention in recent years, where the mainstream approaches are in \emph{pointwise} manners. Meanwhile, \emph{pairwise} manners have shown great potential in supervised metric learning and unsupervised contrastive learning.
Thus, a natural question is raised: does learning in a pairwise manner \emph{mitigate} label noise?
To give an affirmative answer, in this paper, we propose a framework called \emph{Class2Simi}: it transforms data points with noisy \emph{class labels} to data pairs with noisy \emph{similarity labels}, where a similarity label denotes whether a pair shares the class label or not.
Through this transformation, the \emph{reduction of the noise rate} is theoretically guaranteed, and hence it is in principle easier to handle noisy similarity labels.
Amazingly, DNNs that predict the \emph{clean} class labels can be trained from noisy data pairs if they are first pretrained from noisy data points.
Class2Simi is \emph{computationally efficient} because not only this transformation is on-the-fly in mini-batches, but also it just changes loss computation on top of model prediction into a pairwise manner.
Its effectiveness is verified by extensive experiments.
\end{abstract}
\section{Introduction}
It is difficult to label large-scale data accurately. Therefore, datasets with label noise are ubiquitous in the era of big data. However, label noise will degenerate the performance of deep networks, because deep networks will easily overfit label noise \citep{zhang2016understanding}. Almost all existing methods deal with the label noise problem in \textit{pointwise} manners. Namely, these methods use pointwise losses (e.g., cross-entropy loss), and pointwise noise corrections (e.g., sample selection, loss correction, label correction, and others) \citep{xia2021instance, li2019learning,  zhang2018deep, xia2020parts, han2020survey}.

On the other hand, methods employing \textit{pairwise} manners are very prevailing and have made a great success in machine learning, e.g., supervised metric learning and unsupervised contrastive learning \citep{qi2019simple, boudiaf2020unifying, chen2020simple, he2020momentum}, where relationships between data points are exploited. 
% For metric learning, similar and dissimilar data pairs are treated as must-link pairs and cannot-link pairs to optimize the metric. For contrastive learning, two augmented views of any given data points are regarded as positive pairs, and two augmented views of different data points are regarded as negative pairs to learn representations.
% Both metric learning and contrastive learning are in the weakly supervised paradigm. 
Intuitively, the pairwise manners require less pointwise supervision information, i.e., class labels, and might be robust to label noise. In this paper, we naturally ask a question: does learning in a pairwise manner mitigate label noise? This motivates us to introduce a pairwise manner to deal with label noise.
% For metric learning, which learns a distance function over instances, the earlier studies use class labels to optimize the metric \citep{zhang2003parametric, goldberger2004neighbourhood}. More recently, pairwise similarity and dissimilarity constraints are provided as supervision to learn a metric \citep{baghshah2009semi}. \citet{qi2019simple} and \citet{boudiaf2020unifying} utilize pairwise information for deep metric learning, achieving the state-of-the-art results. For contrastive learning, pairwise manners are naturally introduced. Two augmented views of any given data points are regarded as positive pairs, and two augmented views of different data points are regarded as negative pairs. With the new representations learned by contrastive learning, semi-supervised learning methods outperform standard supervised learning methods \citep{chen2020simple}.

Here we propose a noise reduction perspective on handling label noise: \textit{Class2Simi}, i.e., transforming training data with noisy class labels into data pairs with noisy similarity labels. A class label shows the class that an instance belongs to, while a similarity label indicates whether or not two instances belong to the same class. We theoretically prove that through this transformation, the noise rate becomes lower (see Theorem \ref{lower_rate}). This is because, given a data pair, of which if one point has an incorrect class label or even if both points have incorrect class labels, the similarity label could be correct. Moreover, this transformation also reduces a multi-class classification problem into a binary classification problem.
% This is because: (1) a noisy similarity label 1 could be correct if two points from a same class flip into another same class; (2) a noisy similarity label 0 could be correct if two points from two different classes do not flip into one same class (either one flip or two flips). 
In label noise learning, the binary problem is easier to handle and a lower noise rate usually results in higher classification performance \citep{patrini2017making}.

\begin{figure*}[h]
	\vspace{-0pt}
	\centering
	\includegraphics[width=1.0\linewidth]{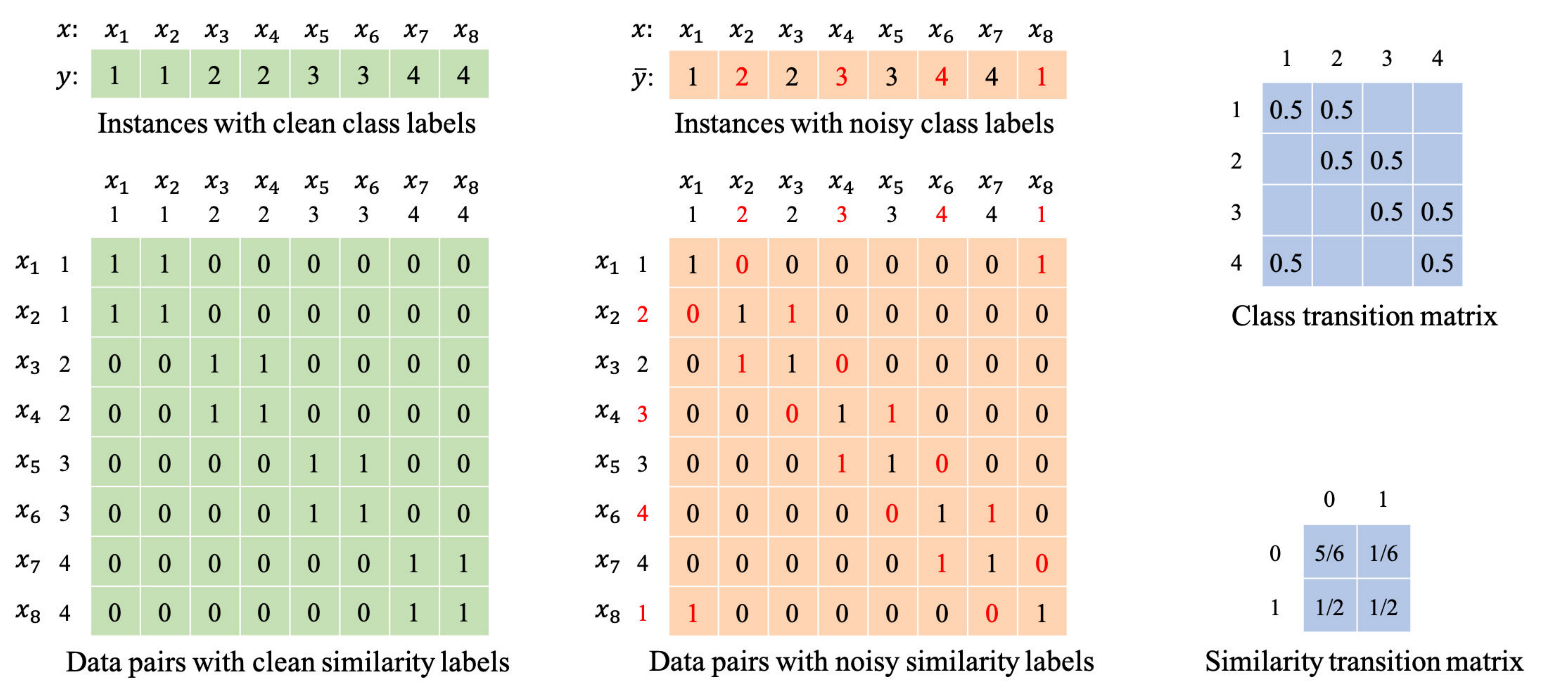}
	\caption{An illustration of the transformation from class labels to similarity labels. Note that $\bar{y}$ stands for the noisy class label and $y$ for the latent clean class label. The labels marked in red are incorrect. If we assume the class label noise is generated according to the transition matrix presented in the upper part of the right column, it can be calculated that the noise rate for the noisy class labels is 0.5 while the noise rate for the noisy similarity labels is 0.25. Note that the transition matrix for similarity labels can be calculated by exploiting the class transition matrix as in Theorem \ref{transform_T}.} \label{fig:label_example}
	\vspace{-10pt}
\end{figure*}

Specifically, we illustrate the transformation and the robustness of similarity labels in Figure \ref{fig:label_example}. In the middle column, we can see the noisy similarity labels of example-pairs $(x_2,x_5)$ and $(x_2,x_4)$ are correct, although there is one mislabeled point in $(x_2,x_5)$, and two mislabeled points in $(x_2,x_4)$. Moreover, if we assume that the noisy class labels in Figure \ref{fig:label_example} are generated according to the latent clean class labels and the class transition matrix (the $ij$-th entry of this matrix denotes the probability that the clean class label $i$ flips into the noisy class label $j$), the noise rate of class labels is $0.5$. Meanwhile, the corresponding similarity transition matrix can be derived from the class transition matrix with the class-priors (see Theorem \ref{transform_T}). The noise rate of similarity labels is $0.25$, which is the proportion of the number of incorrect similarity labels to the number of total similarity labels.

To handle the transformed data pairs with noisy similarity labels, the connection between noisy similarity posterior and clean class posterior should be established. Intuitively, noisy similarity posterior can be linked to clean similarity posterior, and then clean class posterior can be inferred from clean similarity posterior. For the first part, we can draw on the philosophy of dealing with noisy class labels, e.g., selecting reliable data pairs for training, and correcting the similarity loss to learn a robust similarity classifier. For the second part, plenty of similarity metrics can be adopted. As an example, we could adapt the \textit{Forward} \citep{patrini2017making} to learn clean similarity posterior from data with noisy similarity labels. Then, by using the inner product of the clean class posterior \citep{hsu2019multi} to approximate clean similarity posterior, the clean class posterior (and thus the robust classifier) can thereby be learned.  

It is obvious that Class2Simi suffers information loss because we can not recover the class labels from similarity labels, which implies that learning only from similarity labels can only cluster data points but can not identify the semantic classes of clusters. In \citet{hsu2019multi}, a pointwise cluster can be learned from similarity labels. 
% By using a part of the training data with class labels, \citet{hsu2019multi} assigns the dominant class to the output nodes. 
However, in our case, the pairs with similarity labels are constructed from points with class labels, and we could acquire the semantic class information of clusters by pretraining the model from points with class labels without any additional information.
%to initialize the model, so the permutation mentioned above will be the identity map in our case, which means we do not suffer the major information loss in similarity learning. 
% In practice, we first train an auxiliary model from noisy class labels to estimate the class transition matrix, which can learn a rough classier for clean classes. Then we directly initialize the main model with parameters of the auxiliary model without additional training. 
%For a better understanding, we formulate Class2Simi in the form combined with \textit{Forward correction} in following paragraphs. 
% Similarly, the sample selection methods, label correction methods, and other methods can also be adapted to solve the noisy similarity classification problem.
% we first estimate the similarity transition matrix, which bridges the noisy similarity posterior and the clean similarity posterior. The noisy similarity posterior can be learned from the data with noisy similarity labels. Then, given the similarity transition matrix, we can infer the clean similarity posterior from the noisy similarity posterior. Since the clean similarity posterior could be approximated by the inner product of the clean class posterior \citep{hsu2019multi}, the clean class posterior (and thus the robust classifier) can thereby be learned. Similarly, the sample selection methods, label correction methods, and other methods can also be adapted to solve the noisy similarity classification problem.
Note that when class labels of points are corrupted, leading to noisy similarity labels, the proposed pretraining still works because the noisy class is assumed to be dominated by its clean class in label noise learning. Thus we do not suffer the major information loss in noisy similarity learning.

\begin{figure*}[!t]
	\centering
	\includegraphics[width=0.98\textwidth]{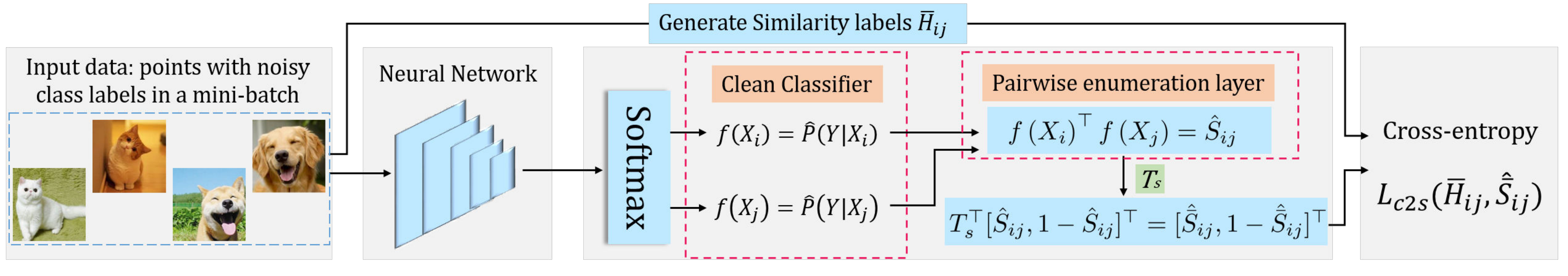}			
	\caption{An overview of the proposed method. We add a pairwise enumeration layer and similarity transition matrix to calculate and correct the predicted similarity posterior. By minimizing the proposed loss $L_{c2s}$, a classifier $f$ can be learned for assigning clean labels. The detailed structures of the Neural Network are provided in Section 4.}
% 	Note that for the noisy similarity labels, some of them are correct and some are not. The similarity label for dogs is correct and the similarity label for cats is incorrect. In practice, the input data is original class-labeled data, and the transformation is conducted during the training procedure rather than before training.
	\label{fig:net}
	\vspace{-10pt}
\end{figure*}

It is worthwhile to mention Class2Simi increases the computation cost very slightly, compared with the standard pointwise training. As shown in Figure \ref{fig:net}, most computation is still pointwise. Only the computation of the pairwise enumeration layer \citep{Hsu18_L2C} and the loss are pairwise, while both the forward and backward propagation are pointwise. The pairwise enumeration layer was verified to only introduce a negligible overhead to the training time \citep{hsu2019multi}. Moreover, the transformation is on-the-fly in mini-batches, which means the pairs are quadratic on the batch size other than the whole sample size.

The contributions of this paper are summarized as follows:
\begin{itemize} [topsep=0pt,itemsep=-1ex,partopsep=1ex,parsep=1ex, leftmargin=0.4cm]
	
	\item We propose a noise reduction perspective on learning with noisy labels, which transforms class labels into similarity labels, reducing the noise rate.
	
	\item We provide a way to estimate the similarity transition matrix $T_s$ by theoretically establishing its relation to the class transition matrix $T_c$. We show even if the $T_c$ is roughly estimated, the induced $T_s$ still works well.
	
	\item We design a deep learning method to learn robust classifiers from data with noisy similarity labels and theoretically analyze its generalization ability.
	
	\item We empirically demonstrate that the proposed method remarkably surpasses the baselines on many datasets with both synthetic noise and real-world noise.
\end{itemize}

The rest of this paper is organized as follows: In Section 2, we formalize the noisy multi-class classification problem. In Section 3, we propose the Class2Simi method and practical implementation. Experimental results are discussed in Section 4. We conclude our paper in Section 5.

%-------------------------------------------------------------------------
\section{Problem Setup and Related Work}
\label{sec:2}
Let $(X,Y)\in\mathcal{X}\times\{1,\ldots,c\}$ be the random variables for instances and clean labels, where $\mathcal{X}$ represents the instance space and $c$ is the number of classes. However, in many real-world applications \citep{zhang2016understanding,zhong2019graph, li2019learning, tanno2019learning, zhang2018deep,xia2021robust,feng2020learning,chou2020unbiased,wu2020multi,zhu2020second, yu2020label, berthon2020confidence}, the clean labels cannot be observed. The observed labels are noisy. Let $\bar{Y}$ be the random variable for the noisy labels. What we have is a sample $\{(x_1,\bar{y}_1),\ldots,(x_n,\bar{y}_n)\}$ drawn from the noisy distribution $\mathcal{D}_\rho$ of the random variables $(X,\bar{Y})$. We aim to learn a robust classifier that could assign clean labels to test data by exploiting the sample with noisy labels.

Existing methods for learning with noisy labels can be divided into two categories: algorithms that result in statistically inconsistent or consistent classifiers.
Methods in the first category usually employ heuristics to reduce the side-effect of noisy labels, e.g., selecting reliable samples \citep{han2018co,yu2019does,Wei_2020_CVPR,wu2020topological,xia2021sample}, reweighting samples \citep{ren2018learning,jiang2018mentornet,ma2018dimensionality,kremer2018robust,reed2014training}, correcting labels \cite{tanaka2018joint,zheng2020error}, designing robust loss functions \cite{zhang2018generalized,xu2019l_dmi,liu2019peer,ma2020normalized}, employing side information \citep{vahdat2017toward,li2017learning}, and (implicitly) adding regularization \citep{li2021provably, li2017learning,veit2017learning,vahdat2017toward,han2018masking,zhang2017mixup,guo2018curriculumnet,hu2020simple,zhang2021learning,han2020sigua}.
Those methods empirically work well in many settings. 
Methods in the second category aim to learn robust classifiers that could converge to the optimal ones defined by using clean data. They utilize the transition matrix, which denotes the probabilities that the clean labels flip into noisy labels, to build consistent algorithms  \citep{natarajan2013learning,scott2015rate,liu2016classification,patrini2017making,northcuttlearning,yu2018learning,kremer2018robust,hendrycks2018using,liu2019peer,yao2020dual,xia2020extended}. 
The idea is that given the noisy class posterior probability and the transition matrix, the clean class posterior probability can be inferred.

Note that the noisy class posterior and the transition matrix can be estimated by exploiting the noisy data, where the transition matrix additionally needs anchor points \citep{liu2016classification,patrini2017making}. 
Some methods assume anchor points have already been given \citep{yu2018learning}. There are also methods showing how to identify anchor points from the noisy training data \citep{liu2016classification}. 

\section{Class2Simi meets noisy supervision}
\label{sec:3}
In this section, we propose a new perspective for learning from noisy data. Our core idea is to transform class labels to similarity labels first, and then handle the noise manifested on similarity labels. 

\subsection{Transformation on labels and the transition matrix}
As in Figure \ref{fig:label_example}, we combine every 2 instances in pairs, and if the two instances have the same class label, we assign this pair a similarity label $1$, otherwise $0$. If the class labels are corrupted, the generated similarity labels also contain noise. 

The definition of the similarity transition matrix is similar to the class one. The elements in a similarity transition matrix denote probabilities that clean similarity labels $H$ flip into noisy similarity labels $\bar{H}$, i.e., $T_{s,mn} \coloneqq P(\bar{H}=n|{H}=m)$. The dimension of the similarity transition matrix is always ${2\times 2}$. Since the similarity labels are generated from class labels, the similarity noise is determined and, thus can be calculated, by the class transition matrix. 

\begin{theorem} 
	\label{transform_T}
	Assume that the dataset is balanced (each class has the same amount of instances, and c classes in total), and the noise is class-dependent. Given a class transition matrix $T_c$, such that $T_{c,ij}=P(\bar{Y}=j|Y=i)$. The elements of the corresponding similarity transition matrix $T_s$ can be calculated as 
	\begin{align*}
		T_{s,00} &= \frac{c^2 - c - \big(\sum_j(\sum_iT_{c,ij})^2-||T_c||_{\mathrm{Fro}}^2\big)}{c^2 - c}, \\
		T_{s,01} &= \frac{\sum_j(\sum_iT_{c,ij})^2-||T_c||_{\mathrm{Fro}}^2}{c^2 - c}, \\
		T_{s,10} &= \frac{c - ||T_c||_{\mathrm{Fro}}^2}{c}, \qquad 
		T_{s,11} = \frac{||T_c||_{\mathrm{Fro}}^2}{c}.
	\end{align*}
\end{theorem}
A detailed proof is provided in Appendix A.

% \begin{proof} \label{proof:th1}
% 	Assume each class has $n$ samples. $n^2T_{c,ij}T_{c,i'j'}$ represents the number of sample-pairs generated by $(\bar{Y}=j|Y=i)$ and $(\bar{Y}=j'|Y=i')$. For the first element $T_{s,00}$, $n^2\sum_{i \neq i'}T_{c,ij}T_{c,i'j'}$ is the number of sample-pairs with clean similarity labels $S = 0$, while $n^2\sum_{i \neq i', j \neq j'}T_{c,ij}T_{c,i'j'}$ is the number of sample-pairs with clean similarity labels $S = 0$ and noisy similarity labels $\bar{S} = 0$. Thus the ratio of these two terms is exact the $T_{s,00}=P(\bar{S}=0|S=0)$. The remaining three elements are obtained in the same way.
% \end{proof}

\begin{remark}
	Theorem \ref{transform_T} can easily extend to the setting where the dataset is unbalanced in classes by multiplying each $T_{c,ij}$ by a coefficient $n_i$. $n_i$ is the number of instances from the $i$-th class.
\end{remark}

Note that the similarity labels are only dependent on class labels. If the class noise is class-dependent, the similarity noise is also `class-dependent' (class means similar and dissimilar). Under class-dependent label noise, a binary classification is learnable as long as $T_{00}+T_{11}>1$ \citep{menon2015learning}, where $T$ is the corresponding binary transition matrix; a multi-class classification is learnable if the corresponding transition matrix $T_c$ is invertible. For Class2Simi, in the most general sense, i.e., $T_c$ is invertible, $T_{s,00}+T_{s,11}>1$ holds. Namely, the learnability of the pointwise classification implies the learnability of the reduced pairwise classification. A proof is provided in Appendix B. However, the latter cannot imply the former: As shown in Figure \ref{fig:label_example}, the class transition matrix is not invertible, and thus the pointwise classification is not learnable while the reduced pairwise classification is learnable. Note that this `learnable' is only for the binary pairwise classification in this case. Technically, two conditions must be met to learn a pointwise classifier from pairwise data: (1) The reduced pairwise classification is learnable; (2) The semantic class information is learnable. Generally, the second condition is equivalent to the learnability of the pointwise classification. Thus the learnability for a pointwise classifier of the two learning manners is consistent.

\begin{theorem} \label{lower_rate}
	Assume that the dataset is balanced (each class has the same amount of samples), and the noise is class-dependent. When the number of classes $c \geq 8$, the noise rate of noisy similarity labels is lower than that of the noisy class labels.
\end{theorem}
A detailed proof is provided in Appendix C.

In multi-class classification problems, the number of classes is usually larger than 8. As $c$ becomes larger, the range of `dissimilarity' of data pairs becomes larger, which is conducive to the reduction of the noise rate. Through Class2Simi, the number of \textit{d-pairs} (with similarity label 0) is $(c - 1)$ times as much as that of \textit{s-pairs} (with similarity label 1). Meanwhile, compared with the original noise rate of noisy class labels, the noise rate of noisy similarity labels of s-pairs is higher and that of d-pairs is lower, while the overall noise rate of data pairs is lower, which partially reflects that the impact of label noise is less bad. Notably, the flip from `dissimilar' to `similar' should be more adversarial and thus more important. In practice, it is common that one class has more than one clusters, while it is rare that two or more classes are in the same cluster. If there is a flip from `similar' to `dissimilar' and based on it we split a (latent) cluster into two (latent) clusters, we still have a high chance to label these two clusters correctly later. If there is a flip from `dissimilar' to `similar' and based on it we join two clusters belonging to two classes into a single cluster, we nearly have zero chance to label this cluster correctly later. As a consequence, the flip from `dissimilar' to `similar' is more adversarial and important, thus deserving a larger weight when calculating the noise rate. Here we assign all data pairs the same weight, otherwise, there would be a more reduction of the noise rate. On balance, considering the reduction of the overall noise rate is meaningful.

When dealing with label noise, a low noise rate has many benefits. The most important one is that the noise-robust algorithms will consistently achieve higher performance when the noise rate is lower \citep{han2018co, xia2019anchor,patrini2017making}. Another benefit is that, when the noise rate is low, the complex instance-dependent label noise can be well approximated by class-dependent label noise \citep{cheng2017learning}, which is easier to handle. 

\begin{figure*}[!t]
	\vspace{-20pt}
	\begin{center}
		\subfigure[\textit{Similar example}]
		{\includegraphics[width=0.46\textwidth]{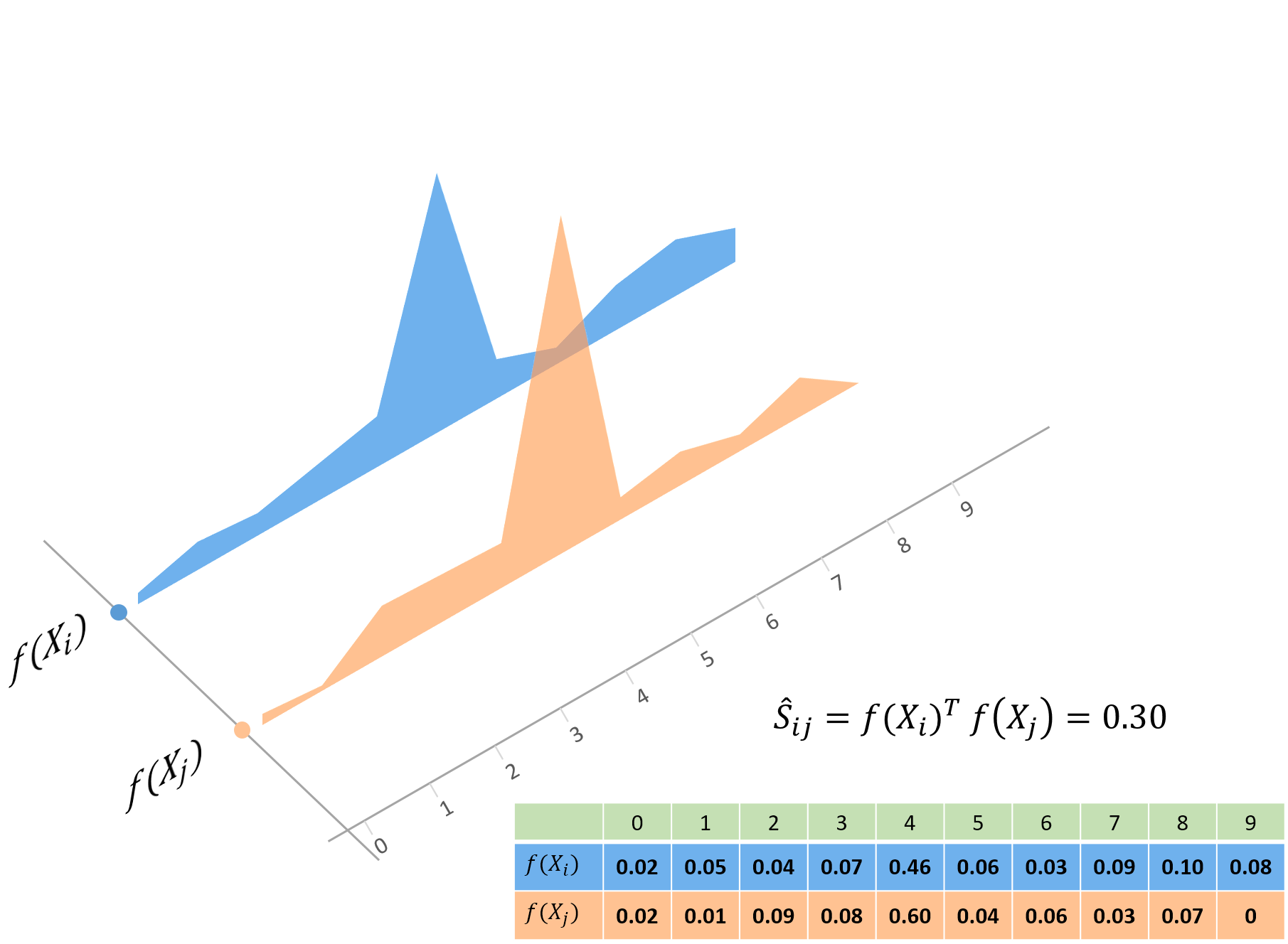}}
		\qquad
		\subfigure[\textit{Dissimilar example}]
		{\includegraphics[width=0.46\textwidth]{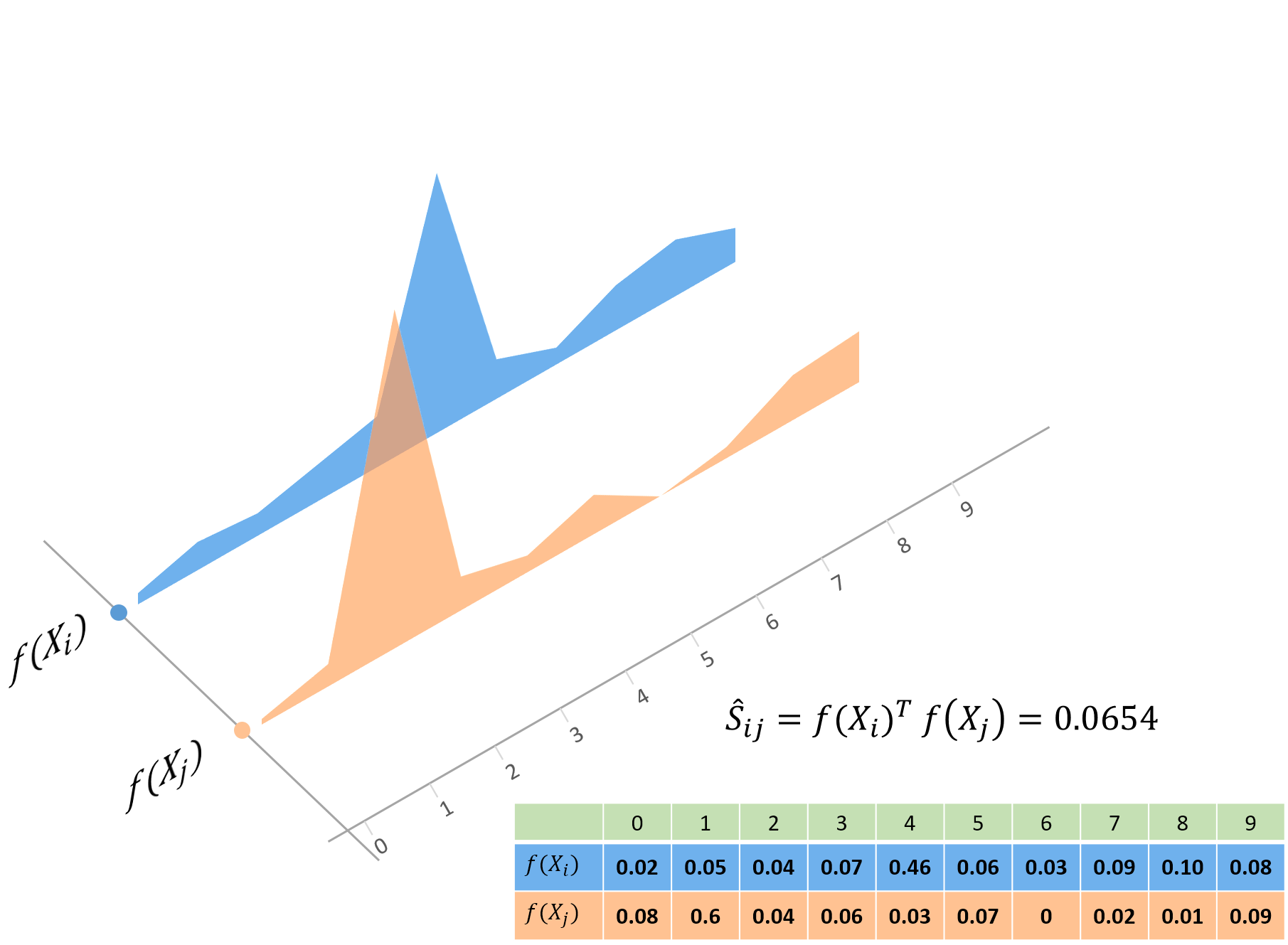}}
	\end{center}
	\caption{Examples of predicted noisy similarity. Assume class number is $10$; $f(X_i)$ and $f(X_{j})$ are categorical distribution of $X_i$ and $X_{j}$ respectively, which are shown above in the form of area charts. $\hat{{S}}_{ij}$ is the predicted similarity posterior between two instances, calculated by the inner product between two categorical distributions.}
	\label{fig:pre_simi}
	\vspace{-8pt}
\end{figure*}

\subsection{Learning with noisy similarity labels}
\label{risk}

In order to learn a multi-class classifier from similarity labeled data, we should establish relationships between class posterior probability and similarity posterior probability. Here we employ the relationship established in \citep{hsu2019multi}, which is derived from a likelihood model. As in Figure \ref{fig:net}, we denote the predicted clean similarity posterior by the inner product between two categorical distributions: $\hat{S}_{i j} = f\left(X_{i}\right)^\top f\left(X_{j}\right)$. Intuitively, $f(X)$ outputs the predicted categorical distribution of input data $X$ and $f(X_i)^\top f(X_{j})$ can measure how similar the two distributions are. For clarity, we visualize the predicted similarity posterior in Figure \ref{fig:pre_simi}. If $X_i$ and $X_{j}$ are predicted belonging to the same class, i.e., $\argmax_{m \in c} f_m(X_i) = \argmax_{n \in c} f_n(X_{j})$, the predicted similarity posterior should be relatively high ($\hat{{S}}_{ij} = 0.30$ in Figure 3(a)). By contrast, if $X_i$ and $X_{j}$ are predicted belonging to different classes, the predicted similarity posterior should be relatively low ($\hat{{S}}_{ij} = 0.0654$ in Figure 3(b)). Note that the noisy similarity posterior $P({\bar{H}}_{ij}|X_i,X_j)$ and clean similarity posterior $P({{H}}_{ij}|X_i,X_j)$ satisfy
\begin{align}
	P({\bar{H}}_{ij}|X_i,X_j) = T_s^{\top}P({{H}}_{ij}|X_i,X_j). \label{eq:n2c}
\end{align}
Therefore, we can infer the predicted noisy similarity posterior $\hat{\bar{S}}_{ij}$ from the predicted clean similarity posterior $\hat{{S}}_{ij}$ with the similarity transition matrix. To measure the error between the predicted noisy similarity posterior $\hat{\bar{S}}_{ij}$ and noisy similarity label $\bar{H}_{ij}$, we employ a binary cross-entropy loss function. The final optimization function is
\begin{align*}
	L_{c2s}&(\bar{H}_{ij},\hat{\bar{S}}_{ij}) \\
	&= - \sum_{i,{j}} \bar{H}_{ij} \log \hat{\bar{S}}_{ij} +   (1-\bar{H}_{ij})\log (1-\hat{\bar{S}}_{ij}).
\end{align*}

The pipeline of the proposed Class2Simi is summarized in Figure \ref{fig:net}. The softmax function outputs an estimation for the clean class posterior, i.e., $f(X)=\hat{P}(Y|X)$, where $\hat{P}(Y|X)$ denotes the estimated class posterior. Then a pairwise enumeration layer is added to calculate the predicted clean similarity posterior $\hat{S}_{ij}$ of every two instances. According to Equation (\ref{eq:n2c}), by pre-multiplying the transpose of the noise similarity transition matrix, we can obtain the predicted noisy similarity posterior $\hat{\bar{S}}_{ij}$. Therefore, by minimizing $L_{c2s}$, we can learn a classifier for predicting noisy similarity labels. Meanwhile, before the transition matrix layer, the pairwise enumeration layer will output a prediction for the clean similarity posterior, which guides $f(X)$ to predict clean class labels.

\begin{remark}
    For a better understanding, we formulate Class2Simi in the form combined with \textit{Forward} as an illustration. However, Class2Simi is a meta method that can be applied on top of sample selection, loss correction, label correction, and many other label noise learning methods. We provide another implementation with \textit{Reweight} in Appendix D.
\end{remark}
\subsection{Implementation}
The proposed algorithm is summarized in Algorithm \ref{alg:c2s}. Since learning only from similarity labels will lose the semantic class information, we load the model trained on the data with noisy class labels to provide the semantic class information for similarity learning in Stage 2.

\subsection{Generalization error}

We formulate the above problem in the traditional risk minimization framework \citep{mohri2018foundations}. The expected and empirical risks of employing estimator $f$ can be defined as
\begin{align*}
	{R}(f) = {E}_{(X_i, X_{j}, \bar{Y}_{i}, \bar{Y}_{j}, \bar{H}_{ij}, T_s)\sim {\mathcal{D}_\rho}}[{\ell}(f(X_i), f(X_{j}), {T}_s, \bar{H}_{ij})],
\end{align*}
and
\begin{align*}
	{R}_n(f) = \frac{1}{n^2}\sum_{i=1}^n\sum_{j=1}^n {\ell}(f(X_i), f(X_{j}),{T}_s, \bar{H}_{ij}),
\end{align*}
where $n$ is the training sample size of the noisy data.
Assume that the neural network has $d$ layers with parameter matrices $W_1,\ldots,W_d$, and the activation functions $\sigma_1,\ldots,\sigma_{d-1}$ are Lipschitz continuous, satisfying $\sigma_{j}(0)=0$. We denote by $H: X\mapsto W_d\sigma_{d-1}(W_{d-1}\sigma_{d-2}(\ldots \sigma_1(W_1X)))\in\mathbb{R}$ the standard form of the neural network. $H =\argmax_{i\in\{1,\ldots,c\}} h_i$. Then the output of the softmax function is defined as $f_i(X)=\exp{(h_i(X))}/\sum_{j=1}^{c}\exp{(h_j(X))}, i=1,\ldots,c$. We can then obtain the following generalization error bound.
\begin{theorem} \label{thm:bound}
	Assume the parameter matrices $W_1,\ldots,W_d$ have Frobenius norm at most $M_1,\ldots, M_d$, and the activation functions are 1-Lipschitz, positive-homogeneous, and applied element-wise (such as the ReLU). Assume the transition matrix is given, and the instances $X$ are upper bounded by $B$, i.e., $\|X\|\leq B$ for all $X$, and the loss function $\ell$ is upper bounded by $M$. Then, for any $\delta>0$, with probability at least $1-\delta$,
	\begin{align}
		{R}&(\hat{f})- {R}_n(\hat{f}) \leq  M\sqrt{\frac{\log{1/\delta}}{2n}} + \nonumber \\
		& \quad  \frac{(T_{s,11} - T_{s,01})2Bc(\sqrt{2d\log2}+1)\Pi_{i=1}^{d}M_i}{T_{s,11}\sqrt{n}}.
	\end{align}
\end{theorem}
A detailed proof are provided in Appendix E.
\begin{algorithm}[t]
	{\bfseries Input}: training data with noisy class labels; validation data with noisy class labels.
	
	\textbf{Stage 1: Learn $\hat{T}_s$}
	
	1: Learn $g(X)=\hat{P}(\bar{Y}|X)$ by training data with noisy class labels, and save the model for Stage 2;

	2: Estimate $\hat{T}_c$ following the optimization method in \citep{patrini2017making};
	
	3: Transform $\hat{T}_c$ to $\hat{T}_s$.
	
	\textbf{Stage 2: Learn the classifier $f(X)=\hat{P}({Y}|X)$}
	
	4: Load the model saved in Stage 1, and train the whole pipeline showed in Figure \ref{fig:net}.
	
	{\bfseries Output}: classifier $f$.
	\caption{Class2Simi}
	\label{alg:c2s}
\end{algorithm}

Theorem \ref{thm:bound} implies that if the training error is small and the training sample size is large, the expected risk ${R}(\hat{f})$ of the representations for noisy similarity posterior will be small. If the transition matrix is well estimated, the clean similarity posterior as well as the classifier for the clean class will also have a small risk according to Equation (\ref{eq:n2c}) and the Class2Simi relations. This theoretically justifies why the proposed method works well. In the experiment section, we will show that the transition matrices are well estimated and that the proposed method significantly outperforms the baselines.

In Class2Simi, a multi-class classification is reduced to a pairwise binary classification. For data pairs, if a surrogate loss is classification-calibrated, minimizing it leads to minimizing the zero-one loss on the pointwise random variables in the limit case. Otherwise, we cannot guarantee the worst-case learnability of learning pointwise labels from pairwise labels, but it cannot imply the average-case non-learnability either. Theoretically, \cite{bao2020similarity} proved that when the pairwise labels are all correct, for the special case $c =2$, a good model for predicting s-/d-pairs must also be a good model for predicting the original classes, under mild assumptions. In practice, it seems fine to use non-classification-calibrated losses. According to \cite{tewari2007consistency}, the multi-class margin loss (i.e., one-vs-rest loss) and the pairwise comparison loss (i.e., one-vs-one loss) are proved to be non-calibrated, but they are still the main multi-class losses in \cite{mohri2018foundations, shalev2014understanding}.

%------------------------------------------------------------------------
\begin{table*}[!t]
\vspace{8pt}
\centering
	\begin{center}
		%		\setcaptionwidth{0.9\textwidth}
		\caption{Means and Standard Deviations of Classification Accuracy over 5 trials on image datasets.} 
		% CE, learned from clean class labels, achieves 99.27 in accuracy with the same notwork and optimizer.}
		\renewcommand{\arraystretch}{1.4} 
		\vspace{0pt}
		\label{tab:syn1}
		\scalebox{0.95}
		{
			\begin{tabular}{c|ccc|ccc}
				\hline
				\textbf{\textit{MNIST}}  & Sym-0.2 & Sym-0.4 & Sym-0.6 & Asym-0.2 & Asym-0.4 & Asym-0.6 \\ \hline
				Co-teaching & 97.34$\pm$0.26 & 94.68$\pm$0.52 & 93.36$\pm$0.47 & 97.37$\pm$0.20 & 96.63$\pm$0.41 & 91.33$\pm$0.38\\
				JoCor & 97.48$\pm$0.12 & 96.31$\pm$0.20 & 93.18$\pm$0.27 & 97.31$\pm$0.09 & 95.73$\pm$0.29 & 91.43$\pm$0.28\\
				PHuber-CE & 98.65$\pm$0.18 & 98.17$\pm$0.15 & 97.63$\pm$0.36 & 98.73$\pm$0.09 & 98.36$\pm$0.25 & 97.37$\pm$0.41 \\
				APL & 98.77$\pm$0.21 & 97.06$\pm$0.37 & 97.67$\pm$0.35 & 98.72$\pm$0.10 & 98.45$\pm$0.29 & 97.58$\pm$0.25\\
				S2E & 98.96$\pm$0.27 & 93.27$\pm$2.18 & 89.37$\pm$0.70 & 99.19$\pm$0.05 & 94.47$\pm$1.08 & 92.36$\pm$2.40\\
				Revision & 98.92$\pm$0.09 & 98.42$\pm$0.50 & 98.10$\pm$0.37 & 98.97$\pm$0.06 & 98.58$\pm$0.19 & 98.21$\pm$0.19 \\ \hline
				Reweight & 98.78$\pm$0.16 & 98.26$\pm$0.22 & 97.02$\pm$0.58 & 98.62$\pm$0.19 & 98.12$\pm$0.31 & 96.98$\pm$0.29  \\ 
				Forward & 98.76$\pm$0.03 & 98.37$\pm$0.25 & 96.89$\pm$0.49 & 98.61$\pm$0.22 & 98.08$\pm$0.33 & 97.43$\pm$0.25  \\ \hline
				R-Class2Simi & {99.04$\pm$0.06} & {98.87$\pm$0.06} & {98.40$\pm$0.17} & {99.06$\pm$0.05} & {98.75$\pm$0.08} & {98.23$\pm$0.20}\\ 
				F-Class2Simi & \textbf{99.26$\pm$0.07} & \textbf{99.18$\pm$0.06} & \textbf{98.91$\pm$0.09} & \textbf{99.26$\pm$0.05} & \textbf{99.08$\pm$0.07} & \textbf{98.91$\pm$0.07}\\ \hline
			    %\hline
				\textbf{\textit{CIFAR10}}   & Sym-0.2 & Sym-0.4 & Sym-0.6 & Asym-0.2 & Asym-0.4 & Asym-0.6 \\ \hline
				Co-teaching & 88.92$\pm$0.45 & 85.97$\pm$1.02 & 75.97$\pm$1.33 & 89.14$\pm$0.36 & 84.77$\pm$1.08 & 76.07$\pm$1.27 \\
				JoCor & 88.46$\pm$0.25 & 85.19$\pm$0.75 & 77.03$\pm$0.92 & 88.96$\pm$0.70 & 85.19$\pm$0.58 & 75.76$\pm$1.31\\
				PHuber-CE & 90.37$\pm$0.26 & 86.05$\pm$0.37 & 74.06$\pm$0.92 & 90.73$\pm$0.22 & 86.06$\pm$0.53 & 73.25$\pm$1.04\\
				APL & 89.07$\pm$0.92 & 85.77$\pm$0.84 & 70.06$\pm$1.06 & 89.97$\pm$0.19 & 85.60$\pm$0.91 & 72.33$\pm$1.68 \\
				S2E & 90.04$\pm$1.22 & 82.05$\pm$1.95 & 57.96$\pm$4.70 & 90.12$\pm$0.97 & 83.16$\pm$1.58 & 64.77$\pm$3.06 \\
				Revision & 90.02$\pm$0.48 & 85.47$\pm$0.71 & 73.92$\pm$2.02 & 89.77$\pm$0.28 & 85.32$\pm$1.36 & 75.24$\pm$1.87  \\ \hline
				Reweight & 89.05$\pm$0.32 & 84.60$\pm$0.45 & 74.87$\pm$1.18 & 89.28$\pm$0.26 & 84.61$\pm$0.62 & 72.77$\pm$1.91 \\ 
				Forward & 89.63$\pm$0.20 & 87.08$\pm$0.31 & 73.24$\pm$1.33 & 90.03$\pm$0.41 & 86.64$\pm$0.71 & 77.41$\pm$0.43  \\ \hline
				R-Class2Simi & 90.91$\pm$0.26 & 87.80$\pm$0.23 & 79.19$\pm$1.65 & 91.07$\pm$0.21 & 87.78$\pm$0.33 & 78.56$\pm$0.63 \\ 
				F-Class2Simi & \textbf{91.38$\pm$0.19} & \textbf{88.22$\pm$0.19} & \textbf{79.45$\pm$0.53} & \textbf{91.24$\pm$0.27} & \textbf{87.79$\pm$0.36} & \textbf{79.05$\pm$0.56} \\ 
				\hline
	
				\textbf{\textit{CIFAR100}}   & Sym-0.2 & Sym-0.4 & Sym-0.6 & Asym-0.2 & Asym-0.4 & Asym-0.6 \\ \hline
				Co-teaching & 57.14$\pm$0.49 & 52.62$\pm$1.03 & 37.32$\pm$1.67 & 57.82$\pm$0.37 & 51.32$\pm$0.83 & 35.32$\pm$1.68 \\
				JoCoR & 58.32$\pm$0.71 & 51.76$\pm$1.07 & 37.02$\pm$1.33 & 58.61$\pm$0.30 & 49.18$\pm$1.05 & 37.09$\pm$1.82\\
				PHuber-CE & 57.90$\pm$0.31 & 52.36$\pm$0.77 & 37.93$\pm$0.86 & 57.33$\pm$0.71 & 51.29$\pm$0.96 & 36.03$\pm$1.34\\
				APL & 54.03$\pm$0.92 & 49.06$\pm$0.93 & 36.06$\pm$2.02 & 55.62$\pm$0.92 & 48.37$\pm$0.94 & 35.02$\pm$1.72\\
				S2E & 59.37$\pm$1.09 & 43.29$\pm$1.94 & 30.08$\pm$3.91 & 58.92$\pm$1.21 & 42.88$\pm$2.16 & 29.93$\pm$4.05 \\
				Revision & 59.62$\pm$0.97 & 53.26$\pm$0.84 & 35.82$\pm$2.06 & 58.77$\pm$0.93 & 52.72$\pm$1.38 & 37.72$\pm$1.75  \\ \hline
				Reweight & 49.59$\pm$0.74 & 39.72$\pm$0.57 & 22.79$\pm$1.35 & 48.87$\pm$0.96 & 36.65$\pm$0.90 & 17.24$\pm$1.97  \\ 
				Forward & 48.68$\pm$0.57 & 39.78$\pm$1.23 & 27.01$\pm$0.89 & 47.90$\pm$0.23 & 37.89$\pm$0.57 & 21.71$\pm$1.53  \\ \hline
				R-Class2Simi & {55.45$\pm$0.55} & {50.38$\pm$0.49} & {35.57$\pm$0.75} & {54.95$\pm$0.65} & {47.56$\pm$0.72} & {34.82$\pm$0.58} \\ 
				F-Class2Simi & \textbf{60.26$\pm$0.18} & \textbf{54.85$\pm$0.60} & \textbf{40.38$\pm$0.58} & \textbf{59.10$\pm$0.13} & \textbf{52.99$\pm$0.78} & \textbf{38.69$\pm$2.84} \\ \hline
			\end{tabular}
		}
	\end{center}
	\vspace{0pt}
\end{table*}

\begin{table*}[!t]
\vspace{8pt}
\centering
	\begin{center}
		%		\setcaptionwidth{0.9\textwidth}
		\caption{Means and Standard Deviations of Classification Accuracy over 5 trials on text datasets.} 
		% CE, learned from clean class labels, achieves 99.27 in accuracy with the same notwork and optimizer.}
		\renewcommand{\arraystretch}{1.4} 
		\vspace{0pt}
		\label{tab:text}
		\scalebox{0.95}
		{
			\begin{tabular}{c|ccc|ccc}
				\hline
				\textbf{\textit{NEWS20}}  & Sym-0.2 & Sym-0.4 & Sym-0.6 & Asym-0.2 & Asym-0.4 & Asym-0.6 \\ \hline
				Co-teaching & 55.32$\pm$0.28 & 51.09$\pm$1.06 & 47.07$\pm$0.83 & 55.29$\pm$0.41 & 53.08$\pm$0.26 & 45.63$\pm$0.75   \\
				JoCor & 52.21$\pm$0.70 & 49.84$\pm$0.92 & 48.83$\pm$0.43 & 55.58$\pm$0.27 & 49.35$\pm$0.62 & 46.21$\pm$0.73\\
				PHuber-CE & 55.73$\pm$0.38 & 54.33$\pm$0.92 & 45.05$\pm$0.49 & 56.76$\pm$0.26 & 51.15$\pm$0.65 & 41.59$\pm$1.05 \\
				APL & 56.91$\pm$0.21 & 53.12$\pm$1.21 & 43.60$\pm$1.28 & 56.11$\pm$0.23 & 50.93$\pm$1.05 & 43.60$\pm$1.28\\
				S2E & 57.93$\pm$0.37 & 47.16$\pm$1.32 & 28.53$\pm$5.04 & 54.89$\pm$1.92 & 50.42$\pm$1.71 & 30.67$\pm$3.12\\
				Revision & 58.06$\pm$0.19 & 52.30$\pm$1.73 & 46.84$\pm$1.09 & 56.41$\pm$0.77 & 53.44$\pm$0.83 & 43.77$\pm$1.08\\ \hline
				Reweight & 53.34$\pm$1.08 & 50.15$\pm$1.33 & 44.73$\pm$0.79 & 53.37$\pm$0.66 & 49.82$\pm$0.44 & 39.46$\pm$1.27  \\ 
				Forward & 57.30$\pm$0.32 & 53.94$\pm$0.42 & 46.91$\pm$1.48 & 53.58$\pm$0.54 & 49.90$\pm$1.44 & 42.55$\pm$3.81  \\ \hline
				R-Class2Simi & \textbf{58.67$\pm$0.38} & 56.59$\pm$0.74 & 50.48$\pm$0.97 & 58.44$\pm$0.66 & \textbf{55.03$\pm$1.55} & \textbf{47.75$\pm$2.17}\\ 
				F-Class2Simi & 58.27$\pm$0.47 & \textbf{56.70$\pm$1.13} & \textbf{50.18$\pm$0.89} & \textbf{58.46$\pm$0.68} & 54.92$\pm$1.66 & 46.07$\pm$3.54\\ \hline
			\end{tabular}
		}
	\end{center}
	\vspace{0pt}
\end{table*}

% \begin{figure*}
% \begin{minipage}[t]{.48\linewidth}
%     \centering
% 	\includegraphics[width=1.0\textwidth]{fig/noise on T.pdf}		
% 	\vspace{-20pt}	
% 	\caption{Means and Standard Deviations of Classification Accuracy over 5 trials on \textit{MNIST}, \textit{CIFAR10} and \textit{CIFAR100} with perturbational ground-truth $\hat{T}_c$.}
% 	\label{fig:tnoise}
% \end{minipage} \quad \ \ \ \ \begin{minipage}[t]{.5\linewidth}
% \vspace{-90pt}
%     \captionsetup{type=table}
%     \caption{Classification Accuracy on \textit{News20}.}
% 	\vspace{-0pt}
% 	\label{tab:text}
% 	\renewcommand{\arraystretch}{1.2}
% 		\begin{tabular}{c|cc|c}
% 			\hline
% 			\textbf{\textit{News20}}  & Sym-0.2 & Sym-0.4 & Asym-0.3 \\
% 			\hline
% 			Forward & 57.30$\pm$0.32 & 53.94$\pm$0.42 & 56.33$\pm$0.61 \\ 
% 			F \& C2S & \textbf{58.27$\pm$0.47} & \textbf{56.70$\pm$1.13} & \textbf{58.20$\pm$0.37} \\ \hline
% 			Reweight & 53.34$\pm$1.08 & 50.15$\pm$1.33 & 52.85$\pm$0.53 \\ 
% 			R \& C2S & \textbf{58.67$\pm$0.38} & \textbf{56.59$\pm$0.74} & \textbf{58.24$\pm$0.40} \\ \hline
% 		\end{tabular}
% \end{minipage}
% \end{figure*}

\section{Experiments}
\textbf{Experiment setup.}\ \
% \textit{MNIST} has $28 \times 28$ grayscale images of 10 classes including 60,000 training images and 10,000 test images. \textit{CIFAR-10} and \textit{CIFAR-100} both have $32 \times 32 \times 3$ color images including 50,000 training images and 10,000 test images. \textit{CIFAR-10} has 10 classes while \textit{CIFAR-100} has 100 classes.
We employ three widely used image datasets, i.e., \textit{MNIST} \citep{lecun1998mnist}, \textit{CIFAR-10}, and \textit{CIFAR-100} \citep{krizhevsky2009learning}, one text dataset \textit{News20}, and one real-world noisy dataset \textit{Clothing1M} \citep{xiao2015learning}. \textit{News20} is a collection of approximately 20,000 newsgroup documents, partitioned nearly evenly across 20 different newsgroups. \textit{Clothing1M} has 1M images with real-world noisy labels and additional 50k, 14k, 10k images with clean labels for training, validation and test, and we only use noisy training set in the training phase. Note that the similarity learning method of Class2Simi is based on \textit{clustering} because there is no class information. 
Intuitively, for a noisy class, if most instances in it belong to another specific class, we can hardly identify it. For example, assume that a class with noisy labels $\bar{i}$ contains $n_i$ instances with ground-truth labels $i$ and $n_j$ instances with ground-truth labels $j$. If $n_j$ is bigger than $n_i$, the model will cluster class $i$ into $j$. Unfortunately, in \textit{Clothing1M}, most instances with label `5' belong to class `3' actually. Therefore, we merge the two classes and denote the modified dataset by \textit{Clothing1M*} which contains 13 classes. For all the datasets, we leave out 10\% of the training data as a validation set, which is for model selection.

For \textit{MNIST}, \textit{CIFAR-10}, and \textit{CIFAR-100}, we use LeNet \citep{lecun1998gradient}, ResNet-26 with shake-shake regularization \citep{Gastaldi17ShakeShake}, and ResNet-56 with pre-activation \citep{he2016identity}, respectively. For \textit{News20}, we first use GloVe \citep{pennington2014glove} to obtain vector representations for the raw text data, and employ a 3-layer MLP with the Softsign active function. For \textit{Clothing1M*}, we use pre-trained ResNet-50 \citep{he2016deep}. Further details for the experiments are provided in Appendix F.1.

% We use the same optimization method as Forward correction to learn the transition matrix $\hat{T_c}$. In Stage 2, we use the Adam optimizer with an initial learning rate 0.001. On \textit{MNIST}, the batch size is 128 and the learning rate decays every 20 epochs by a factor of 0.1 with 60 epochs in total. On \textit{CIFAR-10}, the batch size is also 128 and the learning rate decays every 40 epochs by a factor of 0.1 with 120 epochs in total. On \textit{CIFAR-100}, the batch size is 1000 and the learning rate drops at epoch 80 and 160 by a factor of 0.1 with 200 epochs in total. On \textit{News20}, the batch size is 128 and the learning rate decays every 10 epochs by a factor of 0.1 with 30 epochs in total. On \textit{Clothing1M*}, the batch size is 32 and the learning rate drops every 5 epochs by a factor of 0.1 with 10 epochs in total.

\textbf{Noisy labels generation.}\ \
For clean datasets, we artificially corrupt the class labels of training and validation sets according to the class transition matrix. Specifically, for each instance with clean label $i$, we replace its label by $j$ with a probability of $T_{c,ij}$. In this paper, we consider both symmetric and asymmetric noise settings which are defined in Appendix F.2. \textit{Sym-0.2} means symmetric noise type with a 0.2 noise rate and \textit{Asym-0.2} means asymmetric noise type with a 0.2 noise rate.

\textbf{Baselines.}\ \
In this paper, we compare our method with the following baselines: \textit{Reweight} \citep{liu2016classification}, \textit{Forward} \citep{patrini2017making}, and \textit{Revision} \citep{xia2019anchor}, which utilize a class-dependent transition matrix to model the noise, and learn a robust classifier. Besides, we externally conduct experiments on \textit{Co-teaching} \citep{han2018co}, which is a representative algorithm of selecting reliable samples for training; \textit{JoCoR} \citep{Wei_2020_CVPR}, which employs a joint loss function to select small-loss samples; \textit{PHuber-CE} \citep{menon2020can}, which introduces gradient clipping to mitigate the effects of noise; \textit{APL} \citep{ma2020normalized}, which applies simple normalization on loss functions and makes them robust to noisy labels; \textit{S2E} \citep{yao2020searching}, which properly controls the sample selection process so that deep networks can benefit from the memorization effect. Besides, we conduct experiments on another implementation of the proposed method, which employs \textit{Reweight} (More details are provided in Appendix D). To distinguish these two methods, we call them `F-Class2Simi' and `R-Class2Simi'.

\begin{figure}
    \centering
    \includegraphics[width=\columnwidth]{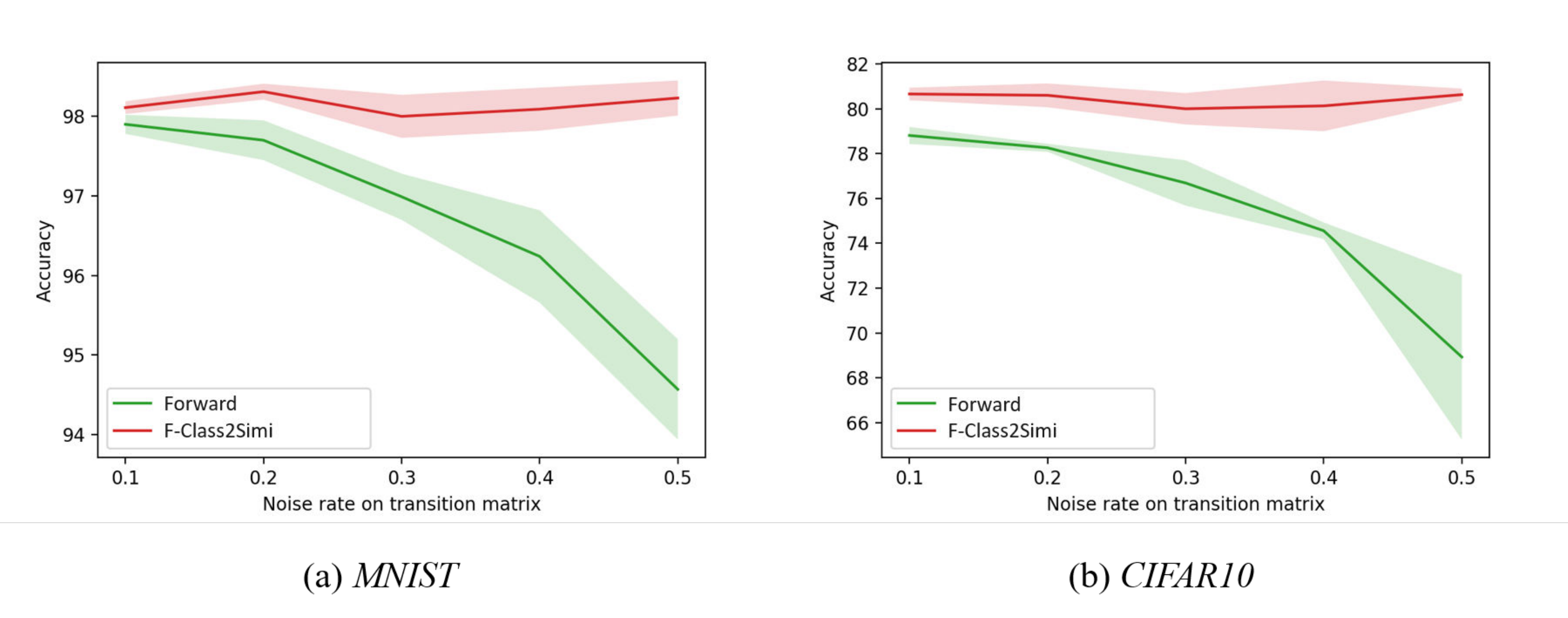}
    \caption{Means and Standard Deviations of Classification Accuracy over 5 trials on \textit{MNIST} and \textit{CIFAR10} with perturbational ground-truth $\hat{T}_c$.}
    \label{fig:tnoise}
\end{figure}

\textbf{Results on noisy image datasets.}\ \ The results in Table \ref{tab:syn1} and Figure \ref{fig:tnoise} demonstrate that Class2Simi achieves distinguished classification accuracy and is robust against the estimation errors on the transition matrix.

From Table \ref{tab:syn1}, overall, we can see that after the transformation, better performance are achieved due to a lower noise rate and the similarity transition matrix being robust to noise. Even for challenging noise rates of 0.6, Class2Simi achieves good accuracy, uplifting about 5 and 10 points on \textit{CIFAR-10} and \textit{CIFAR-100} respectively, compared with the corresponding pointwise methods.

In Figure \ref{fig:tnoise}, we show that the similarity transition matrix is robust against estimation errors. To verify this, we add some random noise to the ground-truth $T_c$ through multiplying every element in class $T_c$ by a random variable $\alpha_{ij}$. We control the noise rate on the $T_c$ by sampling $\alpha_{ij}$ in different intervals, i.e., 0.1 noise means that $\alpha_{ij}$ is uniformly sampled from $\pm[1.1,1.2]$. Then we normalize $T_c$ to make its row sums equal to 1. From Figure \ref{fig:tnoise}, we can see that the accuracy of Forward drops dramatically with the increase of the noise on $T_c$. By contrast, there is only a slight fluctuation of F-Class2Simi, indicating Class2Simi is robust against the estimation errors on the transition matrix.

%------------------------------------------------------------------------

% \begin{table}[t]
% \centering
% \vspace{0pt}
% 	\caption{Classification Accuracy on \textit{News20}.}
% 	\vspace{-0pt}
% 	\label{tab:text}
% 	\renewcommand{\arraystretch}{1.4}
% 	\scalebox{.9} 
% 	{
% 		\begin{tabular}{c|cc|c}
% 			\hline
% 			\textbf{\textit{News20}}  & Sym-0.2 & Sym-0.4 & Asym-0.3 \\
% 			\hline
% 			Co-teaching & 55.32$\pm$0.28 & 51.09$\pm$1.06 & 54.78$\pm$0.95 \\
% 			JoCoR & 52.21$\pm$0.70 & 49.84$\pm$0.92 & 49.88$\pm$0.64\\
% 			PHuber-CE & 55.73$\pm$0.38 & 54.33$\pm$0.92 & 55.62$\pm$0.70\\
% 			APL & 56.91$\pm$0.64 & 53.12$\pm$1.21 & 55.62$\pm$0.86\\
% 			S2E & 57.93$\pm$0.37 & 47.16$\pm$1.32 & 53.76$\pm$1.63 \\
% 			Revision & 58.06$\pm$0.19 & 52.30$\pm$1.73 & 55.72$\pm$0.49\\  \hline
% 			Reweight & 53.34$\pm$1.08 & 50.15$\pm$1.33 & 52.85$\pm$0.53 \\ 
% 			Forward & 57.30$\pm$0.32 & 53.94$\pm$0.42 & 56.33$\pm$0.61 \\  \hline
% 			R-Class2Simi & \textbf{58.67$\pm$0.38} & {56.59$\pm$0.74} & \textbf{58.24$\pm$0.40} \\
% 			F-Class2Simi & {58.27$\pm$0.47} & \textbf{56.70$\pm$1.13} & {58.20$\pm$0.37} \\ \hline
% 		\end{tabular}
% 	}
% \end{table}
\vspace{0pt}

%\vspace{15pt}

\begin{table}[t]
\vspace{0pt}
\centering
	\caption{Classification Accuracy on \textit{Clothing1M*}. }
	%KCL, MCL and MNS only have access to noisy similarity labels.}
	%CE, using original class labels for training, achieves 67.04 in accuracy with the same notwork and optimizer. }
	\label{tab:real1}
	\renewcommand{\arraystretch}{1.4} 
	\scalebox{1.0}
	{
		\begin{tabular}{c|c|c|c}
			\hline
			Co-teaching & 74.70 & JoCoR & 74.98 \\\hline 
			PHuber-CE & 73.16 & APL & 58.93 \\\hline
			S2E & 72.30 & Revision & 74.65 \\ \hline
			Forward & 73.88 & F-Class2Simi & 75.41\\ \hline
			Reweight & 74.44 & R-Class2Simi & \textbf{75.76}
			\\ \hline
		\end{tabular}
	}
\end{table}	

\begin{table}[!h]
\vspace{0pt}
\centering
	\caption{Classification Accuracy on clean datasets. CE uses class labels and the cross-entropy loss function. C2S refers to Class2Simi.}
	\vspace{-0pt}
	\label{tab:abaltion}
	\renewcommand{\arraystretch}{1.4} 
	\scalebox{0.8} 
	{
		\begin{tabular}{c|c|c|c|c}
			\hline
			Dataset & \textbf{\textit{MNIST}} & \textbf{\textit{CIFAR10}} & \textbf{\textit{CIFAR100}} & \textbf{\textit{News20}}\\				\hline
			CE & \textbf{99.30$\pm$0.02} & 94.03$\pm$0.14  &58.74$\pm$0.51& \textbf{59.86$\pm$0.39} \\
			\hline
			C2S & 99.24$\pm$0.05 & \textbf{94.05$\pm$0.27} &\textbf{60.36$\pm$0.89}& 59.74$\pm$0.20\\ \hline
		\end{tabular}
	}
	\vspace{-8pt}
\end{table}

\textbf{Results on the noisy text dataset.}\ \ Results in Table \ref{tab:text} show that the proposed method works well on the text dataset under both symmetric and asymmetric noise settings.

\textbf{Results on the real-world noisy dataset.}\ \ Results in Table \ref{tab:real1} show that the proposed method also performs well against agnostic noise.

\textbf{Ablation study.}\ \ To investigate how the similarity loss function influences the classification accuracy, we conduct experiments with the cross-entropy loss function and the similarity loss function on clean datasets over 3 trials, where the $T_c$ is set to an identity matrix. All other settings are kept the same. As shown in Table \ref{tab:abaltion}, on \textit{MNIST}, \textit{CIFAR10}, and \textit{News20}, the similarity loss function does not improve the classification accuracy on clean data, and on \textit{CIFAR100}, the improvement is marginal. However, in Table \ref{tab:syn1} and \ref{tab:text}, the improvements are significant, which reflects the improvements are mainly benefited from the lower noise rate and the reduced noisy binary paradigm.

%------------------------------------------------------------------------
\section{Conclusion}
This paper proposes a noise reduction perspective on dealing with class label noise by transforming training data with noisy class labels into data pairs with noisy similarity labels. We establish the connection between noisy similarity posterior and clean class posterior and propose a deep learning framework to learn classifiers from the transformed noisy similarity labels. The core idea is to transform pointwise information into pairwise information, which makes the noise rate lower. We also prove that not only the similarity labels but the similarity transition matrix is robust to noise. Experiments are conducted on benchmark datasets, demonstrating the effectiveness of our method. In future work, investigating different types of noise for diverse real-life scenarios might prove important.

%------------------------------------------------------------------------
\section*{Acknowledgments}
SHW, XBX, and TLL were supported by Australian Research Council Project DE-190101473. BH was supported by the RGC Early Career Scheme No. 22200720, NSFC Young Scientists Fund No. 62006202 and HKBU CSD Departmental Incentive Grant. NNW was supported by National Natural Science Foundation of China Grant 61922066, Grant 61876142. GN and MS were supported by JST AIP Acceleration Research Grant Number
JPMJCR20U3, Japan. MS was also supported by Institute for AI and Beyond, UTokyo.

%\newpage

\bibliography{main}
\bibliographystyle{icml2021}

\end{document}